# The Future of Scientific Simulations: from Artificial Life to Artificial Cosmogenesis.

Clément Vidal
Evolution, Complexity and Cognition research group
Vrije Universiteit Brussel (Free University of Brussels)
Krijgskundestraat 33, 1160 Brussels, Belgium
Phone +32-2-640 67 37
Fax +32-2-644 07 44
http://clement.vidal.philosophons.com
clement.vidal@philosophons.com

**Abstract**:
This philosophical paper explores the relation between modern scientific simulations and the future of the universe. We argue that a simulation of an entire universe will result from future scientific activity. This requires us to tackle the challenge of simulating open-ended evolution at all levels in a single simulation. The simulation should encompass not only biological evolution, but also physical evolution (a level below) and cultural evolution (a level above). The simulation would allow us to probe what would happen if we would "replay the tape of the universe" with the same or different laws and initial conditions. We also distinguish between *real-world* and *artificial-world* modelling. Assuming that intelligent life could indeed simulate an entire universe, this leads to two tentative hypotheses. Some authors have argued that we may already be in a simulation run by an intelligent entity. Or, if such a simulation could be made real, this would lead to the production of a new universe. This last direction is argued with a careful speculative philosophical approach, emphasizing the imperative to find a solution to the heat death problem in cosmology. The reader is invited to consult Annex 1 for an overview of the logical structure of this paper.

## Content

*the laws of the universe have engineered their own comprehension*

*(Davies 1999, 146)*

# Introduction

What will happen to the Earth and the Sun in the far future? The future story depicted by modern science is a gloomy one. In about 6 billion years, it will be the end of our solar system, with our Sun turning into a red giant star, making the surface of Earth much too hot for the continuation of life as we know it. The solution then appears to be easy: move. However, even if life would colonize other solar systems, there will be a progressive end of all stars in galaxies. Once stars have converted the available supply of hydrogen into heavier elements, new star formation will come to an end. In fact, the problem is worse. It is estimated that even very massive objects such as black holes will evaporate in about $10^{98}$ years (Adams and Laughlin 1997).

This scenario is commonly known as the "heat death", and says that the universe will irreversibly decay towards a state of maximum entropy **[b, d]**[1]. If this model is correct **[c]**, then it clearly means that the indefinite continuation of life is impossible in this universe **[f]**. What is the point of living in a universe doomed to annihilation? Ultimately, why should we try to solve mundane challenges of our daily lives and societies, if we can not even imagine a promising future for intelligent life in the universe? If we recognize this heat death **[1.12]**, then we should certainly do something to avoid it **[1.13]**, and thus try to change the future of the universe **[1.14]**.

A few authors have proposed some speculative solutions, but we'll see that they are insufficient because none of them presently allows the indefinite continuation of intelligent life. We will instead argue that intelligent civilization will in the far future produce a new universe **[4.0]**. Although it sounds like a surprising proposition, resembling science fiction scenarios, we will consider it seriously and carefully.

It should be noted that the proposition of involving intelligent life into the fate of the universe is at odds with traditional science. Indeed, the modern scientific worldview has often suggested that the emergence of intelligence was an accident in a universe that is completely indifferent to human concerns, goals, and values (e.g. Weinberg 1993; Stenger 2007). I thus challenge this proposition, and another one that is commonly associated with it, which says that: **[a]** intelligent civilization can not have a significant influence on cosmic evolution.

Our central focus is on the future of scientific simulations, and how important this activity could be in the far future, if intelligent civilization is to have influence on cosmic evolution. It is increasingly clear that simulations and computing resources are becoming main tools of scientific activity **[1.15]**. More concretely, at a smaller scale than the universe, we have already begun to produce and "play" with artificial worlds, with the practice of computer simulations. In particular, efforts in the Artificial Life (ALife) research field have shown that it is possible to create digital worlds with their own rules, depicting agents evolving in a complex manner. We will see that such simulation promise to become more and more complex and elaborated in the future.

1 Letters and numbers in bold and brackets refer to the two maps in Annex 1. Please refer to this annex for more details.

In the first part, we argue that the path towards a simulation of an entire universe is an expected outcome of our scientific simulation endeavours. We then examine how such a simulation could be realized (instantiated, made physical) and solve the irreversible heat death of the universe, expected to happen at some future time.

# Towards a simulation of an entire universe

In this section, we argue that simulating open-ended evolution not only in biology, but also to link it to physical evolution (a level below) and to cultural evolution (a level above) will be a long-term outcome of our scientific simulation endeavours. Such a simulation would allow us to probe what would happen if we would "replay the tape of the universe". We then discuss in more depth the status and potential usefulness of a simulation of an entire universe, making a distinction between *real-world* and *artificial-world* modelling. We outline and criticize the "simulation hypothesis", according to which our universe has been proposed to be just a simulation. Let us first summarize the historical trend of exponential increase of computing resources.

## *Increase of computing resources*

We may note two important transitions in the history of human culture. The first is the *externalization of memory* through the invention of writing. This allowed an accurate reproduction and a safeguard for knowledge. Indeed, knowledge could easily be lost and distorted in an oral tradition. The second is the *externalization of computation* through the invention of computing devices. The general purpose computer was inspired by the work of Church, Gödel, Kleene and Turing, and its formal specifications constitute the most general computing device (see Davis 2000 for a history of computation). The consequences of this last transition are arguably as significant -or even more significant- as the invention of writing. In particular, the changes induced by the introduction of computers in scientific inquiry are important, and remain underestimated and understudied (see however (Floridi 2003) for a good starting point).

Computing resources have grown exponentially, at least for over a century. There is much literature about this subject (see e.g. Kurzweil 1999; 2006). Moore's "law" famously states that the number of transistors doubles every 18 months on a single microprocessor **[1.21]**. Exponential increase in processing speed and memory capacity are direct consequences of the law. What are the limits of computer simulations in the future? Although there is no Moore's law for the efficiency of our algorithms, the steady growth in raw computational power provides free "computational energy" to increase the complexity of our simulations. This should lead to longer term and more precise predictions. Apart from the computational limitation theorems (uncomputability, the computational version of Gödel's theorem proved by Turing), the only limit to this trend is the physical limit of matter or the universe itself (Lloyd 2000; Krauss and Starkman 2004). As argued by Lloyd (2000; 2005) and Kurzweil (2006, 362) it should be noted that the ultimate computing device an intelligent civilization could use in the distant future is a very dense object, i.e. a black hole **[1.22]**.

From a cosmic outlook, Moore's trend is in fact part of a much more general trend which started with the birth of galaxies. The cosmologist and complexity theorist Eric Chaisson proposed a quantitative metric to characterize the dynamic (not structural) complexity of physical, biological and cultural complex systems (Chaisson 2001; 2003). It is the *free energy rate density* (noted $\Phi_M$) which is the rate at which free energy transits in a complex system of a given mass (Fig. 1). Its dimension is energy per time per mass (erg $s^{-1}$ $g^{-1}$). Let us illustrate it with some examples (Chaisson 2003, 96). A star has a value ~1, planets ~$10^2$, plants ~$10^3$, humans ~$10^4$ and their brain ~$10^5$, current microprocessors ~$10^{10}$. According to this metric, complexity has risen at a rate faster than exponential in recent times **[1.20]**. We might add along this complexity increase, the hypothesis that there is a tendency to do ever more, requiring ever less energy, time and space; a phenomenon also called *ephemeralization* (Fuller 1969; Heylighen 2007), or "Space-Time Energy Matter" (STEM) compression (Smart 2008). This means that complex systems are increasingly localized in space, accelerated in time, and dense in energy and matter flows.

In Tomas Ray's simulation *Tierra* (Ray 1991), digital life competes for CPU time, which is analogous to energy in the organic world. The analogue of memory is the spatial resource. The agents thus compete for fundamental properties of computers (CPU time, memory) analogous to fundamental physical properties of our universe. This design is certainly one of the key reasons for the impressive growth of complexity observed in this simulation.

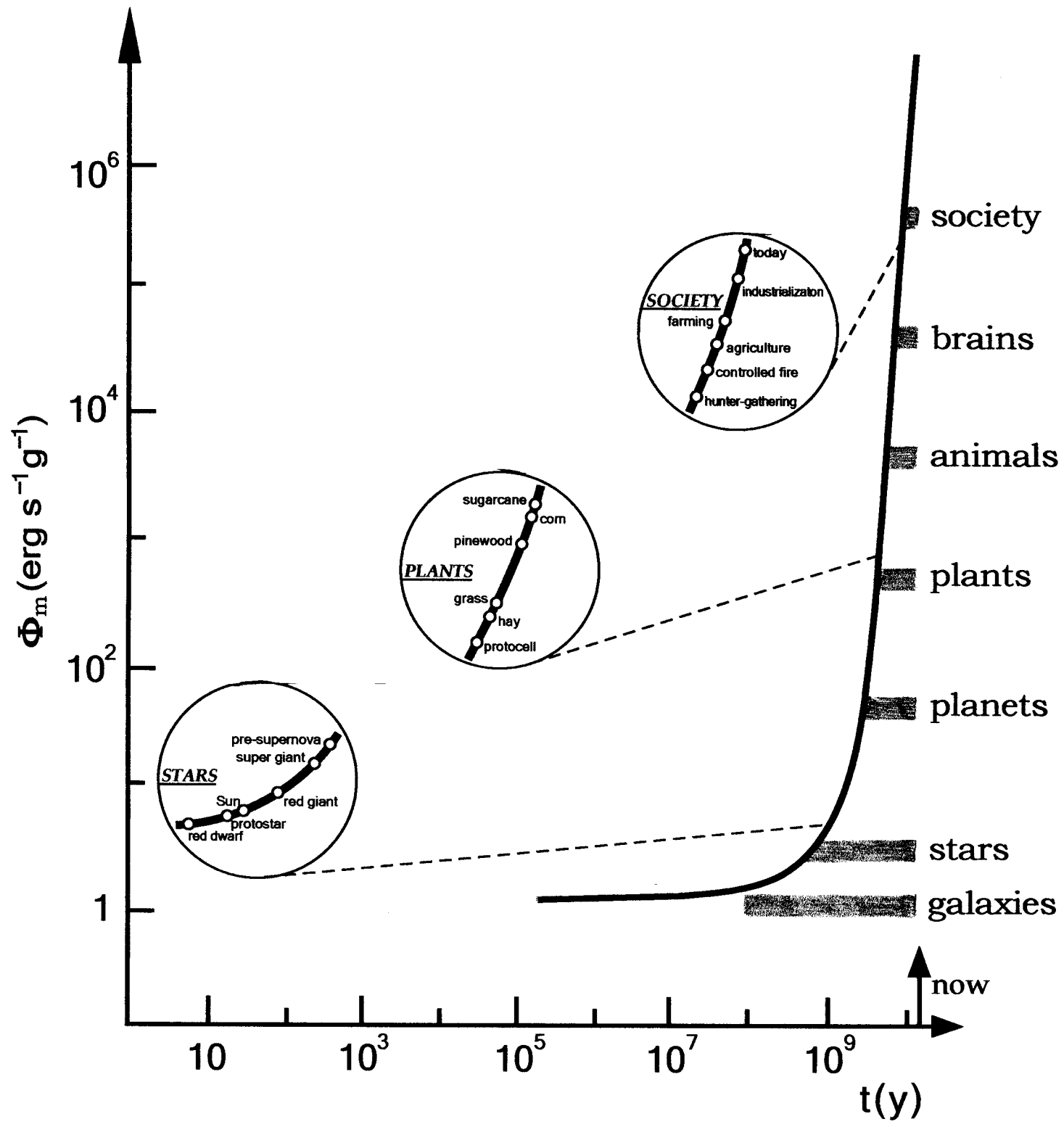

Fig. 1. Figure excerpted from (Chaisson 2003, 97). The original caption is: "The rise of free energy rate density, $\Phi_M$, plotted as histograms starting at those times when various open structures emerged in Nature, has been rapid in the last few billion years, much as expected from both subjective intuition and objective thermodynamics. The solid curve approximates the increase in normalized energy flows best characterizing the order, form and structure for a range of systems throughout the history of the Universe. The circled insets show greater detail of further measurements or calculations of the free energy rate density for three representative systems - stars, plants and society - typifying physical, biological and cultural evolution, respectively. Many more measures are found in Chaisson (2001)." Note that microprocessors are outside the scale of this diagram since they appear at $10^{10}$ on the $\Phi_M$ axis.

## *Bridging physical, biological and cultural evolution*

We saw that a metric can be found to compare complex systems traditionally considered as different in nature. This important insight is just a first step towards bridging physical, biological and cultural evolution **[1.32]**. The information-theoretic endeavours are certainly going in this direction (e.g. (Von Baeyer 2004; Prokopenko, Boschetti, and Ryan 2007; Gershenson 2007; Floridi 2003) as well as "Big History" thinkers (e.g. Christian 2004; Spier 2005).

Artificial Life (ALife) is a field of research examining systems related to life, its processes, and its evolution through simulations using either computer models (soft ALife), robotics (strong ALife), or biochemistry (wet ALife). A general challenge for ALife is to obtain an artificial system capable of generating open-ended evolution (Bedau et al. 2000). Some results have been obtained linking for example the evolution of language with quasi-biological traits (Steels and Belpaeme 2005). Working towards the design of a digital universe simulating the rise of levels of complexity in the physical, biological and cultural realms is the challenge of simulating an entire universe **[1.16]**. An important step in this direction, although it stays on the physical level, is the "Millennium Run" simulation , which starts from the very beginning of the universe to generate the large scale structures of the universe (Springel et al. 2005).

However, we must acknowledge important difficulties of conceptual, methodological and cultural integration between the different disciplines involved. In such an endeavour, human-made social and academic boundaries between disciplines of knowledge must be overcome **[1.31]**. I proposed to construct integrative scientific worldviews (or philosophies) with *systems theory*, *problem solving* and *evolutionary theory* as three generic interdisciplinary approaches (Vidal 2008). There should be a seamless link between simulations in physics, biology and social sciences (culture). If this would happen, we would have the basic tools to work towards a model and a simulation of the entire universe **[1.33; 2.0]**. In fact the search for such bridges is obviously necessary if we want to tackle such difficult problems as the origin of life, where we aim to explain the emergence of life out of physico-chemical processes.

## *Replaying the tape of the universe*

The biologist Stephen Jay Gould (1990) asked the famous question: "what would remain the same if the tape of life were replayed?". Paraphrasing and extending it to the universe, the question becomes: "what would remain the same if the tape of the universe were replayed?". We should first notice that the tape metaphor has its limits. Indeed, if the tape and its player were perfect, we should get exactly the same results when re-running the tape. Yet if our universe self-constructs, one question is whether small fluctuations could lead to slightly different outcomes, or very different ones if for example the system is chaotic.

By exploring other simulated universes, this approach would allow us to face one of the main difficulties in cosmology, which is that, as far as we know, there is only one object of study: our unique universe. More precisely, two fundamental limitations of current cosmology that Ellis (2005, sec. 3) has pointed out might then be addressed:

> **Thesis A1: The universe itself cannot be subjected to physical experimentation**. *We cannot re-run the universe with the same or altered conditions to see what would happen if they were different, so we cannot carry out scientific experiments on the universe itself.* Furthermore,

**Thesis A2: The universe cannot be observationally compared with other universes.** *We cannot compare the universe with any similar object, nor can we test our hypotheses about it by observations determining statistical properties of a known class of physically existing universes.*

Interestingly, re-running the tape of the universe is also a very relevant research program for tackling the difficult "fine-tuning" problem in cosmology, which states that if any of a number of parameters, fundamental constants in physics and initial conditions in cosmology were slightly different, no complexity of any sort would come into existence (see e.g. (Leslie 1989) for a good review). To give just one example of fine tuning, let us consider the ratio of electrical and gravitational forces between protons, which is $10^{36}$. Changes either in electromagnetism or in gravity « by only one part in $10^{40}$ would spell catastrophe for stars like the sun » (Davies 1984, 242).

Victor Stenger (1995; 2000) has performed a remarkable simulation of possible universes. He considered four fundamental constants, and then analysed "100 universes in which the values of the four parameters were generated randomly from a range five orders of magnitude above to five orders of magnitude below their values in our universe, that is, over a total range of ten orders of magnitude" (Stenger 2000). Anthony Aguirre did a similar work by exploring classes of cosmologies with different parameters (Aguirre 2001). These simulations are only an early attempt in simulating other possible universes, and the enterprise is certainly worth pursuing, with more complex models, more parameters to vary, etc.

The simulation of an entire universe can be seen as perhaps the ultimate challenge of simulations in science. But what kind of simulation would it be? What could it be used for? To answer these questions we will now distinguish between two kinds of modelling.

## *Real-world and artificial-world modelling*

A computer simulation can be defined as a model where some aspects of the world are chosen to be modelled and the rest ignored. When in turn such a simplified model is run on hardware that is significantly more computationally efficient than the physical system being modelled, this makes it possible to run the model faster than the phenomena modelled, and thus to make predictions of our world. The paradigm of Artificial Life (ALife) strongly differs from traditional modelling, by studying not only "life-as-we-know-it", but also "life-as-it-could-be" (Langton 1992, sec. 1). We propose to extend this modelling technique to any *process* and not just to life, leading to the more general distinction of *processes-as-we know-them* and *processes-as-they-could-be* (Red'ko 1999) . We call the two kinds of modelling respectively *real-world* modelling and *artificial-world* modelling.

*Real-world* modelling is the endeavour to model *processes-as-we-know-them*. This includes traditional scientific modelling, such as models in physics, weather forecast models, but also applied evolutionary models, etc. The goal of such models is to better understand our world, and make predictions about it. For what would a *real-world* simulation of an entire universe be useful? At first glance, it would provide us very good understanding of and predictive power over our world. However, this view has some severe limitations. First, if the simulation is really of the entire universe, it should be "without anything left out". This is a strange situation, since it would imply that the model (simulation) is as complex as our universe. Such a simulation would thus not provide a way to systematically predict all aspects of our universe, because it would not be possible to run it faster than real physical processes. Another limiting argument is that more computational power does not necessarily mean better predictive abilities. This is pretty clear when considering chaotic systems such as the weather, which rapidly become unpredictable. A simulation still has to be simpler than reality if it is to be of any practical use. This means that in the context of "replaying the tape of our universe", we would still have to investigate a simplified simulation of our universe.

*Artificial-world* modelling is the endeavour to model *processes-as-they-could-be*. The formal fundamental rules of the system (of life in the case of ALife) are sought. The goal of ALife is not to model life exactly as we know it, but to decipher the most simple and general principles underlying life and to implement them in a simulation. With this approach, one can explore new, different life-like systems. Stephen Wolfram (2002) has a similar approach by exploring different rules and initial conditions on cellular automata, and observing the resulting behaviour of the system. It is legitimate to emphasize that this is a "new kind of science". Indeed, this is in sharp contrast with traditional science focusing on modelling or simulating reality. There is thus a creative aspect in the artificial-world modelling, which is why many artists have enthusiastically depicted imaginary ALife worlds. For what would an artificial-world simulation of an entire universe be useful? We would be able not only to "replay the tape of *our* universe", but also to play and replay the tape of *other possible* universes (thus tackling limitations A1 and A2 explicated by Ellis) **[2.1; 2.2]**. We saw that simulation constitutes a research program for tackling the fine-tuning issue in cosmology **[2.3]**. The concept of "a universe" then needs to be redefined and extended, since we only know by definition our unique universe.

Should this artificial world modelling of an entire universe be interpreted as a *simulation* or as a *realization* (Pattee 1989)? To start, let us consider the first possibility, with the *simulation hypothesis.*

## *The simulation hypothesis*

Let us assume what we have argued in the previous section, i.e. that intelligent life will indeed be able at some point to simulate an entire universe. If such a simulation is purely digital, thus pursuing the research program of soft ALife, this leads to the *simulation hypothesis*, which has two main aspects. First, looking into the future, it means that we would effectively create a whole universe simulation, as has been imagined in science fiction stories and novels such as the ones of Isaac Asimov (1956) or Greg Egan (2002). Very well then! A second possibility is that we ourselves could be part of a simulation run by a superior intelligence (e.g. Bostrom 2003; Barrow 2007; Martin 2006). Although these scenarios are fascinating, they suffer from two fundamentals problems. First, the "hardware problem" : on what physical device would such a simulation run? Is there an infinity of simulation levels? Second, such an hypothesis violates Leibniz' logical principle of the identity of the indiscernibles. Leibniz' principle states that "if, for every property F, object $x$ has F if and only if object $y$ has F, then $x$ is identical to $y$". Let $x$ be reality, and $y$ be the supposed simulated universe we would be living in. If we have no way to distinguish between them, they are identical. Unless we find a "bug" in reality, or a property F that could only exist in a simulation and not in reality, this hypothesis seems useless. A more comprehensive criticism of these discussions can be found in (McCabe 2005).

The ontological status of this simulation would be reflected by the states of the hardware running it, whatever the realistic nature of the simulation. From this point of view, we can argue that it remains a *simulation*, and not a *realization* (Harnad 1994). Is there another possibility for realizing the simulation of an entire universe? That is what we will explore now.

# Towards a realization of an entire universe

We first outline some aspects concerning the far future of the universe. We then put forward a philosophical approach to tackle this problem, and outline a speculative solution called "artificial cosmogenesis".

## *The heat death problem*

We outlined in introduction the heat death problem. Consider the second law of thermodynamics which is one of the most general laws of physics. It states that the entropy of an isolated system will tend to increase over time. Hermann von Helmholtz applied it to the universe as a whole in 1854 to state the heat death (HD) problem, i.e. that the universe will irreversibly go towards a state of maximum entropy. Modern cosmology shows that there are some other models of the end of the universe (such as Big Bounce, Big Rip, Big Crunch..., see (Vaas 2006) for an up-to-date review). The point is that none of them allows the possibility of the indefinite continuation of life as we know it. The study of the end state of the universe, or *physical eschatology*, is a scattered but exciting field of research that we cannot detail more here (see (Ćirković 2003) for an extensive literature guide).

Some speculative scenarios have been proposed to tackle this problem. They all suppose as we do in this paper that "intelligent civilization *can* have significant influence on cosmic evolution" **[4.1]**; but also that in the future, life will be very different as the one we know. Let us mention some of them. Dyson proposed that life and communication can continue "forever", utilizing a finite store of energy (Dyson 1979); the "final anthropic principle" put forward by Barrow and Tipler (1986) proposes that intelligent information-processing will never die out. Interestingly, under certain conditions, it is theoretically possible to make computing a reversible process (Bennett 1982; Landauer 1991; Krauss and Starkman 2000). If we could make this happen, this might be a way to possibly have "life" continue for an indefinite amount of time.

These speculations are remarkable in the sense that they attempt to find ways for intelligent life to survive forever. However, they assume the additional hypothesis that life should take another "information-like" form. Krauss and Strakman (2000) showed that there are serious difficulties to the scenario proposed by Dyson. The reversible computation scenario is also not sustainable in the long run, since, as Krauss and Strakman argue, no finite system can perform an infinite number of computations with a finite amount of energy. Furthermore, these scenarios give no clear link to the increasing abilities of intelligent life to model the universe, nor do they relate to the fine-tuning problem.

In an optimistic picture, that is if our civilization does not self-destruct (or if it does, we can add the hypothesis that we are not alone in the universe...), we can see the HD problem as the longest-term problem for intelligent life in the universe. How should we react to it? Charles Darwin's thought on the HD problem remains perfectly relevant: "Believing as I do that man in the distant future will be a far more perfect creature than he now is, it is an intolerable thought that he and all other sentient beings are doomed to complete annihilation after such long-continued slow progress" (Darwin 1887, 70)

## *A philosophical approach for a speculative topic*

Before proposing another possible solution to the HD problem, we have to make a methodological clarification. The solution proposed in the next section will be approached from a *speculative* philosophical stance, as opposed to *critical* philosophy (Broad 1924). I have proposed a general philosophical framework to rationally construct speculative theories in (Vidal 2007). We should be well aware of the difficulty of the question we are tackling; an age-old philosophical problem which is: "what is the ultimate fate of humanity and the universe in the very distant future?". This problem is philosophical because (1) we do not have unambiguous empirical or experimental support to favour a unique outcome and (2) it is such an ambitious question, that the proposed answer can only be tentative and speculative. It is however still very worth considering because the philosophical inquiry aims to advance our most profound questions here and now, whatever their difficulty and our limited knowledge.

## *Artificial cosmogenesis*

The cosmologist Lee Smolin proposed a theory called Cosmological Natural Selection (CNS) in order to tackle the fine-tuning problem (Smolin 1992; 1997). According to this natural selection of universes theory, black holes give birth to new universes by producing the equivalent of a Big Bang, which produces a baby universe with slightly different physical properties (constants, laws). This introduces variation, while the differential success in self-reproduction of universes via their black holes provides the equivalent of natural selection. This leads to a Darwinian evolution of universes whose properties are fine tuned for black hole generation, a prediction that can in principle be falsified.

Smolin is not the only cosmologist reasoning with multiple universes, comprising an extended ensemble called a multiverse. Although the idea of a multiverse is a speculative one, it is increasingly popular among many cosmologists. New universes are generally theorized to appear from the inside of black holes, or from the Big Bang itself **[3.0; 3.1]**. Kuhn (2007) distinguished many kinds of multiverse models: by disconnected regions (spatial); by cycles (temporal); by sequential selection (temporal); by string theory (with minuscule extra dimensions); by large extra dimensions; by quantum branching or selection; by mathematics and even by all possibilities, whatever this may mean. Among these multiverse theories, Smolin's CNS is arguably the most scientifically testable (Smolin 2007).

It should be noted however that in Smolin's theory, (1) the roles of life and intelligence in the universe are incidental, as they are in the modern scientific worldview. Which is, let us remember, the main assumption we challenge here. Another problem is that (2) the theory does not propose a specific mechanism for the variation of universe parameters beyond the assumption of randomness. Is it possible to overcome these two shortcomings? A few authors have dared to extend CNS by including intelligent life into this picture, correcting those two problems and also bringing indirectly a possible solution to the HD problem (Crane 1994; Harrison 1995; Baláz 2005; Gardner 2000; 2003; Smart 2008). Simply stated, the thesis is that advanced intelligent civilization will solve the HD problem by reproducing the universe. This direction can be seen as the ultimate challenge of strong/wet ALife, to realize a new universe.

Let us note however that there is not (yet) a uniform terminology among the five mentioned authors. Inspired by Smolin's terminology we could speak of a "Cosmological *Artificial* Selection" (CAS), artificial selection on simulated universes enhancing natural selection of real universes (Barrow 2001, 151). The biological analogy is interesting here. Humans who practice artificial selection on animals do not "design" or "create" new organisms, nor do they replace natural selection. They just try to foster some traits over others. In CNS, many generations of universes are needed to randomly generate a fine tuned complex universe. In CAS, the extensive simulations prior to the replication event would presumably help greatly to generate a fine tuned universe, that is robust to complexity emergence.

We do not attempt here to go into the details of how a CAS could become realized. However, let us first remember that black holes might be the ultimate computing device for an intelligent civilization **[1.22]**, and that they constitute a possible gateway for the emergence of a new universe in some multiverse theories **[3.1]**.

Along with ALife, constituting a general biology, Pattee (1989) suggested to also consider a general physics. As in ALife, this "Artificial Cosmogenesis" discipline would have two parts. One focusing on "software" universe simulations using computer models (analogous to soft ALife); the other focusing on implementing the software in reality (analogous to strong/wet ALife). It it clear however that the analogue of soft ALife (universe simulation) is only in its infancy, and the analogue of strong/wet ALife (universe realization) lies in the far future.

This solution to the HD problem gives a general challenge to intelligence in the universe: to continue to explore and understand the functioning of our universe so as to possibly reproduce it in the far future **[2.3; 4.0]**. This would make the indefinite continuation of life possible, yet in another universe **[4.2]**. This scenario aslo fits with the ultimate goal of evolution as a whole: survival. It is likely to be a difficult and stimulating enough challenge to encourage and occupy intelligent civilization for the foreseeable future.

The degree of control that intelligence could have in this process still has to be discovered. For example, how much might the physical properties of our existing universe (physics of black holes, etc.) constrain the realization of a new universe? Furthermore, the issue of the ethical responsibility of humanity in this proposition is outside the scope of this paper and remains to be explored (see however (Gardner 2003, Part 6) and (Smart 2008) for two different viewpoints).

# Conclusion

The use of scientific simulations has constituted a revolution in the way we practice science. We have outlined the fast-moving changes occurring in our universe, and argued that the limit of scientific simulations is the simulation of an entire universe. Furthermore, we have formulated an hypothesis that the heat death of complexity in our universe could be avoided through an *artificial cosmogenesis,* a discipline analogous to artificial life.

Scientific inquiry today undertakes to understand our world; in the future, this will be increasingly aided by simulations of our and other possible universes. Such simulations would be indispensable tools if intelligent civilization moves towards an artificial cosmogenesis.

# Annex 1 - Logical structure of the paper.

This annex presents the logical structure of the main arguments presented in this paper represented by two maps. The problem is mapped in Fig. 2. and the proposed solution in Fig. 3. For an easier back-and-forth between the paper and the maps, the blocks are numbered in the map (letters for Fig. 2, and numbers for Fig. 3) and those numbers appear in bold in the text.

This approach provides an *externalization of reasoning* so that arguments can be clearly visualized. This brings many benefits, such as:

- Allowing the reader to quickly and clearly grasp the logic of the argumentation.
- Presenting an alternative structure of the content of the paper. The table of content and the abstract tend to present a rhetorical (and not logical) structure.
- Allowing the possibility of a constructive discussion of assumptions and deductions. For example, a critique can say "the core problem is not P but Q"; or "I disagree that hypothesis [X.XX] leads to [Y.YY], you need implicit hypothesis Z, ..." or "hypothesis [Z.ZZ] is wrong because"; or “there is another solution to your problem, which is...” etc.

It should be clear however that reading those maps can not replace the reading of the paper. Only the core reasoning is mapped, sometimes even in a simplified way.

To draw those maps we used some of the insights of Eliyahu Goldratt's Theory of Constraints (TOC) and its "Thinking Process" (see Goldratt and Cox 1984; Goldratt Institute 2001; Scheinkopf 1999). The TOC is a well proven management technique widely used in finance, distribution, project management, people management, strategy, sales and marketing . We see it and use it as part of a generic problem solving toolbox, where causes and effects are mapped in a transparent way. In our paper, the core problem is: “how to make the indefinite continuation of life possible?”; and the proposed solution is that “intelligent civilization can reproduce the universe”.
In this TOC framework, three fundamental questions are employed to tackle a problem:

(1) What to change?

A core problem is identified as the *undesirable effect*, and mapped in a "Current Reality Tree" (CRT), see Fig. 2.

(2) To what to change?

A solution is proposed and mapped in a "Future Reality Tree" (FRT), which leads to the *desirable effect*, see Fig. 3.

(3) How to cause the change?

A plan is developed to change from CRT to FRT. This third step in the context of this paper is even more speculative, so it is almost not developed, and thus not mapped.

To tackle the problem in practice, six important questions should be addressed, constituting the "six layers of resistance to change". These questions can be used to trigger discussions (Goldratt Institute 2001, 6):

(1) Has the right problem been identified?
(2) Is this solution leading us in the right direction?
(3) Will the solution really solve the problems?
(4) What could go wrong with the solution? Are there any negative side-effects?
(5) Is this solution implementable?
(6) Are we all really up to this?

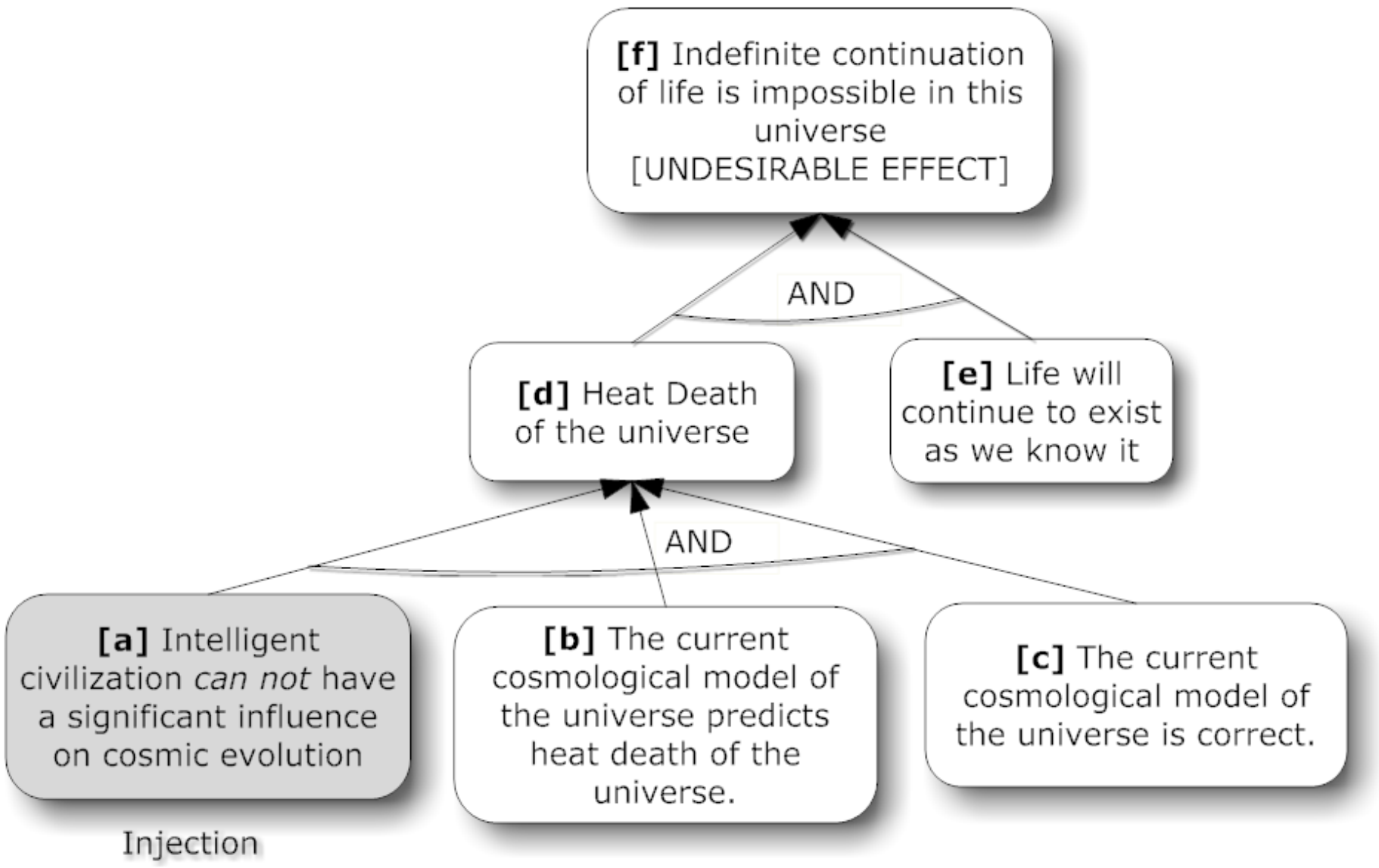

Fig. 2. The Current Reality Tree (CRT) represents the core problem underneath this paper (how to make the indefinite continuation of life possible?). The "injection" (grayed) is the proposition which is challenged. It is the statement that **[a]**: "intelligent civilization can not have any significant influence on cosmic evolution".

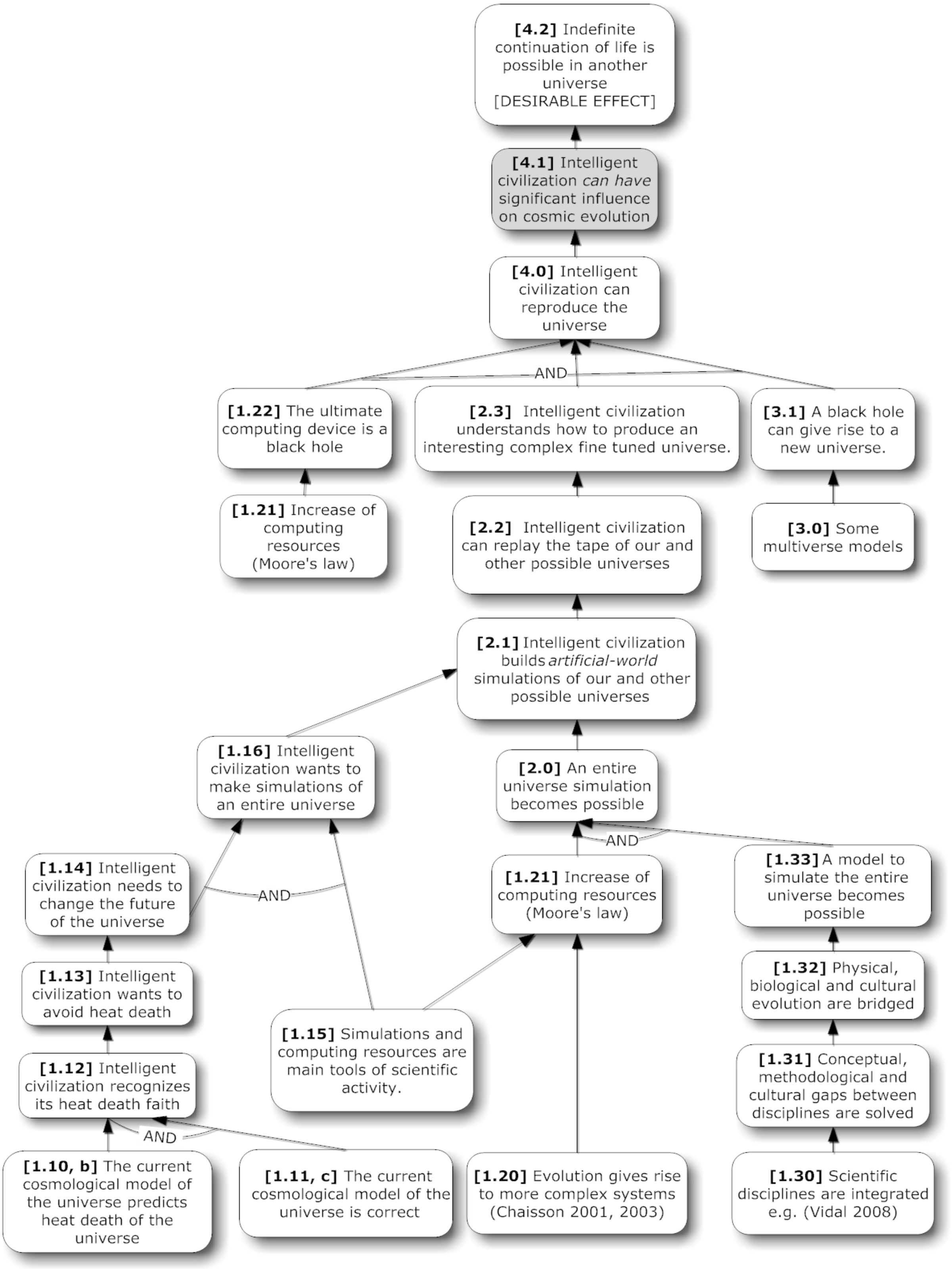

Fig. 3. The Future Reality Tree (FRT) shows the proposed solution to the problem mapped in the CRT (Fig. 2). The diagram can be read by increasing numerical order. The "injection" chosen to solve the core problem is **[4.1]**:"intelligent civilization can have significant influence on cosmic evolution".

## Acknowledgements

I thank three anonymous referees for their sharp comments, my colleagues Milan Ćirković, Carlos Gershenson, Francis Heylighen, Mark Martin, Marko Rodriguez and John Smart for their rich and valuable feedback. I thank Eric Chaisson for his comments and his kind permission to use his curve (Fig. 1). I especially thank Piet Holbrouck for introducing me to the theory of constraints, which gave rise to the annex of this paper. Researchers interested in the topics of this paper (and others) are welcome to join the "Evo Devo Universe" research community at www.evodevouniverse.com.